\documentclass{ifacconf}

\usepackage{graphicx}      
\usepackage{amssymb}
\usepackage{amsmath}
\newtheorem{example}{Example}
\usepackage{amssymb}
\usepackage{xcolor}
\usepackage{natbib}        
\usepackage{multirow}
\usepackage{caption}              
\usepackage{subcaption}           
\usepackage{booktabs}
\usepackage{soul}

\usepackage{todonotes}
\usepackage{combelow}
\usepackage{bm}
\begin{document}
\begin{frontmatter}

\title{Gaussian Process Regression for \\Inverse Problems in Linear PDEs} 

\author[First]{Xin Li} 
\author[Third]{Markus Lange-Hegermann}
\author[First]{Bogdan Rai\cb{t}\u{a}}

\address[First]{Georgetown University, 
   Washington, D.C. 20057 USA \\(e-mail: xl572@georgetown.edu, br607@georgetown.edu).}
\address[Third]{Institute Industrial IT, OWL University of Applied Sciences and Arts, Lemgo, Germany (e-mail: markus.lange-hegermann@th-owl.de).}

\begin{abstract}                
This paper introduces a computationally efficient algorithm in systems theory for solving inverse problems governed by linear partial differential equations (PDEs). We model solutions of linear PDEs using Gaussian processes  with priors defined based on advanced commutative algebra and algebraic analysis. The implementation of these priors is algorithmic and achieved using the Macaulay2 computer algebra software. An example application includes identifying the wave speed from noisy data for classical wave equations, which are widely used in physics. The method achieves high accuracy while enhancing computational efficiency.
\end{abstract}


\end{frontmatter}

\section{Introduction}
Inverse/parameter identification problems governed by linear partial differential equations (PDEs) as described by \cite{bai2024gaussian} arise in numerous scientific and engineering applications, including wave propagation in earthquake simulations as shown by \cite{puel2022mixed} and medical imaging analysis shown by \cite{mang2018pde}. \cite{kaipio2006statistical} state that the goal of an inverse problem is to determine unknown parameters, which often involves addressing challenges such as sensitivity to initial condition and measurement noise. A robust and efficient solution framework is important for obtaining accurate and reliable results in these applications. \cite{Stuart_2010} provides a comprehensive review of Bayesian approaches for inverse problems, offering insights into regularization.  

Gaussian Processes (GPs) provide a powerful probabilistic framework for modeling functions in general and more specifically solutions of PDEs by incorporating prior knowledge and handling noise in data, see \cite{rasmussen2006gaussian}. Recently, \cite{bai2024gaussian} provided a comprehensive discussion on how GPs can be used to model solution of PDEs with a focus on Bayesian approaches. Earlier works such as \cite{RAISSI2017683} explore GPs to infer parameters of differential equations from noisy observations. 

Our approach builds on \citep{lange2018algorithmic,besginow2022constraining,harkonen2023gaussian} by using the Ehrenpreis--Palamodov theorem  in algebraic analysis  to construct suitable GP priors that generate exact solutions of given PDEs. 
The proposed method utilizes the computer algebra software package Macaulay2 to implement the priors algorithmically, enabling efficient computation. Our method can be implemented to any system of linear PDEs and is demonstrated in the example of the 2$d$-wave equation. The results highlight the effectiveness of the proposed approach in reconstructing the wave propagation and learning the unknown wave speed. 

The major contributions of this paper are:
\begin{enumerate}
    \item We develop a general framework for constructing GP priors to model solutions of linear PDEs with constant coefficients, addressing inverse problems.
    \item For noisy data, our method demonstrates the capability to achieve accurate results without requiring additional complexity.
    \item We illustrate that the method achieves a balance between computational efficiency and predictive accuracy, making it a robust choice for resource-constrained environments.
\end{enumerate}

\begin{figure*}[t] 
    \centering
    \begin{subfigure}[b]{0.24\textwidth} 
        \includegraphics[width=\textwidth]{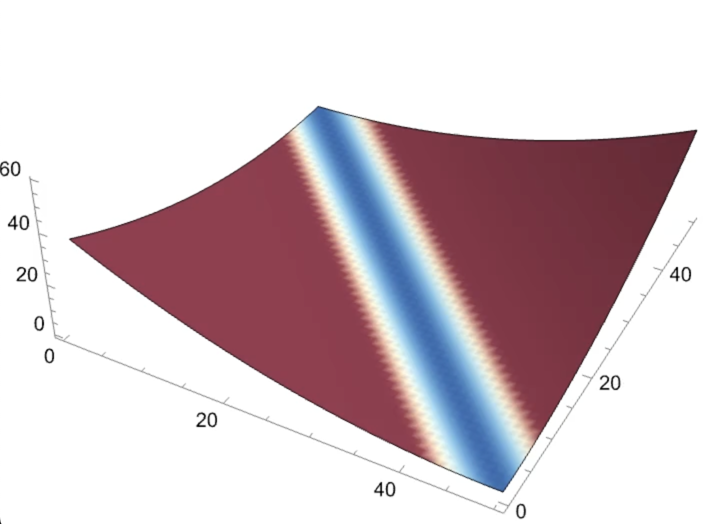}
        \caption{Prediction at $t=0$}
        \label{fig:image11}
    \end{subfigure}
    \hfill
    \begin{subfigure}[b]{0.24\textwidth}
        \includegraphics[width=\textwidth]{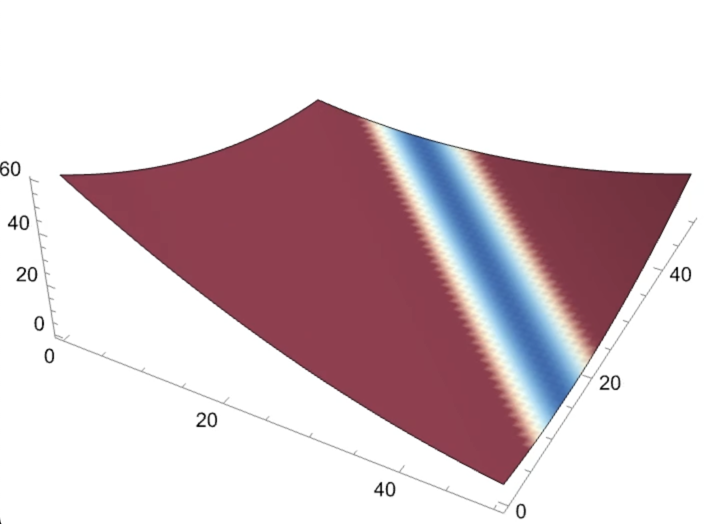}
        \caption{Prediction at $t=1$}
        \label{fig:image22}
    \end{subfigure}
    \hfill
    \begin{subfigure}[b]{0.24\textwidth}
        \includegraphics[width=\textwidth]{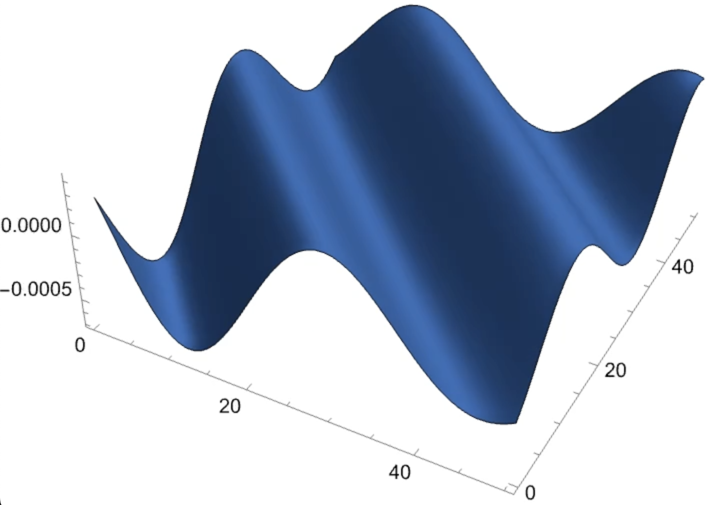}
        \caption{Error for Direct Problem}
        \label{fig:image33}
    \end{subfigure}
    \hfill
    \begin{subfigure}[b]{0.24\textwidth}
        \includegraphics[width=\textwidth]{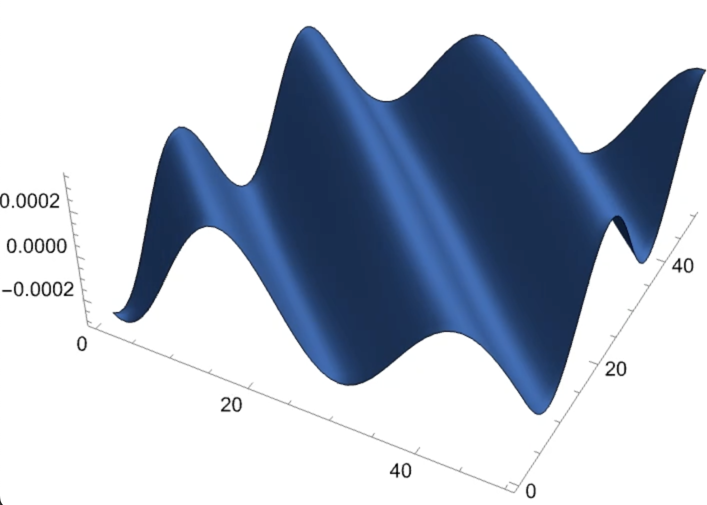}
        \caption{Error for Inverse Problem}
        \label{fig:image44}
    \end{subfigure}
    \caption{Fig. (\ref{fig:image11}) and Fig. (\ref{fig:image22}) show the prediction of our method for the true solution $u(x,y,t) = (x + y - \sqrt{3}t)^2$ of the 2$d$-wave equation $u_{tt} = a^2(u_{xx}+u_{yy})$ at time $t=0$ and $t=1$, demonstrating the expected movement along the diagonal direction of the $xy$-plane. Fig. (\ref{fig:image33}) and Fig. (\ref{fig:image44}) show the difference between the prediction and the true solution for direct problem (known $a = \sqrt{1.5}$) and inverse problem (unknown $a$) respectively at time $t = 0.5$. The experiments are done with 1000 data points $(x_i,y_i,t_i,Y_i)$ where $Y_i=u(x_i,y_i,t_i)$. This very clean data gives us a very good baseline to check how accurate our predictions are. Both predictions perform very well, with errors of the order of $10^{-4}$. It is remarkable  that we solve the inverse problem with similar accuracy to the direct problem, since there is a crucial extra parameter to learn, parameter which appears in the equation directly.
}
    \label{fig:four_images2}
\end{figure*}

\section{Problem Setup and Preliminary} 
\subsection{Ehrenpreis--Palamodov theorem} \label{epsection}
Given the ODE $\frac{d^2y}{dx^2} - y = 0$, there are two distinct roots of the characteristic polynomial $r_1 = 1$ and $r_2 = -1$, so the general solution is $y = c_1 e^{x} + c_2 e^{-x}$ for arbitrary $c_1$ and $c_2$. In particular, solving such differential equations is based on finding zeros of polynomial expressions. For partial differential equations (PDEs), the situation is different. 
+

Instead of a finite set of roots, the solutions are determined by so-called characteristic varieties. For the example of 1$d$-wave equation, $u_{tt} = a^2u_{xx}$, where $a$ represents the wave propagation speed, the characteristic variety consists of the lines $x = \pm at$, which leads to the general solution $u(x,t) = f(x - at) + g(x + at)$ given by d'Alembert's formula. To obtain a general solution representation for PDEs, a key result is the Ehrenpreis--Palamodov theorem, which connects  commutative algebra, functional analysis, and differential equations. Here we state the Ehrenpreis--Palamodov theorem in the case of a single PDE:

\begin{thm}[\cite{palamodov1970linear}]\label{thm:EP}
Let $\Omega\subset\mathbb R^d$ be convex,  $A\in \mathbb{C}[\partial_1, \ldots, \partial_d]$, and $V = \{ z \in \mathbb{C}^d: A(z) = 0 \}$. Then there exist irreducible varieties $V_i$ with $V=V_1\cup\ldots V_r$
and polynomials
$D_{i,j}\in \mathbb{C}[x,z]$ such that 
all smooth solutions $u\in C^\infty(\Omega)$ of $A(\partial)u(x) = 0$ in $\Omega$ can be approximated by
\begin{equation} \label{epequation}
  \sum_{k=1}^m \sum^{r}_{i=1}\sum_{j=1}^{r_i} w_{i,j,k}D_{i,j}(x,z_{i,j,k})e^{\langle x,z_{i,j,k}\rangle} \text{ for }z_{i,j,k}\in V_i.
\end{equation}
\end{thm}
Here multiplication of polynomial variables $\partial_i$ is identified with the action of differentiation on the module $C^\infty(\Omega)$.

Theorem \ref{thm:EP} is relevant to linear PDEs with constant coefficients and states that solutions can be approximated by a superpositions of exponential-polynomial solutions. The polynomials $\{D_{i,j}(x,z)\}$ in \eqref{epequation} are called \textit{Noetherian multipliers}. The Noetherian multipliers $D_{i,j}$ and the varieties $V_i$ can be computed algebraically using Macaulay2 package \texttt{NotherianOperators} under the command \texttt{solvePDE}. See \cite{chen2023noetherian} for details.


\subsection{Characteristic varieties}
The characteristic variety of an operator is an algebraic variety defined as the zero set of the principal operator.
\begin{example} (No PDE)
Let $A$ be an operator and suppose $A=0$. In this case, the characteristic variety includes all points in the complex plane.
\end{example}
\begin{example} (Transport Equation with Constant Coefficients)
Let $L$ be the differential operator for the transport equation with constant coefficients
$
    L = \partial_t - a\partial_x
$
where $a$ is a constant, and represents the speed of propagation. The characteristic variety is the zero set of the  symbol map
$
    \xi_t - a\xi_x = 0
$
which describes a line.
\end{example}
In this paper, the main experiment is based on 2$d$-wave equation with constant coefficients, though the method can be implemented for \textit{any linear PDE with constant coefficients}. The differential operator for the equation is 
\begin{align} \label{2dwaveeq}
    L = \frac{\partial^2}{\partial t^2} - a^2\left(\frac{\partial^2}{\partial x^2} + \frac{\partial^2}{\partial y^2}\right)
\end{align}
where  $a>0$ is an unknown constant which represents the wave speed. The characteristic variety is the set where the  symbol map vanishes:
\begin{align} \label{2dwavecv}
    \xi_t^2 = a^2(\xi_x^2 + \xi_y^2).
\end{align}
This is the equation of a double cone with its vertex at the origin, and the time variable $\xi_t$ is related to the spatial variables $\xi_x$ and $\xi_y$.


\subsection{Gaussian processes}

Following \cite{rasmussen2006gaussian}, a Gaussian Process (GP) $g \sim GP(\mu,k)$ defines a probability distribution on the evaluations of functions $\Omega \rightarrow \mathbb{R}^l$, where $\Omega \subset \mathbb{R}^d$, such that function values $[g(x_1), \ldots , g(x_n)]$ at any points $x_1, \ldots , x_n \in \Omega$ follows a multivariate Gaussian distribution according to the mean function $\mu\colon \Omega \rightarrow \mathbb{R}^l$: 
\begin{equation*}
    \mu\colon x
\mapsto E(g(x))
\end{equation*} and covariance kernel $k\colon\Omega \times \Omega \rightarrow \mathbb{R}^{l\times l}$: 
\begin{align*}
    k\colon (x,x') \mapsto E\big((g(x) - \mu(x))(g(x') - \mu(x'))^T\big)
\end{align*}
GPs are widely used in statistics for modeling and regression, particularly suitable for handling problems with little data. A GP remains a GP under the application of a linear operator, as stated in the following:
\begin{lem}
\label{lemma-1}
Let $g \sim GP(\mu(x), k(x,x'))$ with realizations in the function space $\mathcal{F}^N$, $\mathcal{F} = C^{\infty}(\Omega)$, and $\operatorname{B}: \mathcal{F}^N \rightarrow \mathcal{F}^M$ a  continuous linear operator. 
Then, the pushforward $\operatorname{B}_*g$ of $g$ under $\operatorname{B}$ is a GP with 
\begin{equation*}
    \operatorname{B}_*g \sim GP(\operatorname{B} \mu(x), \operatorname{B} k(x,x')(\operatorname{B}')^T),
\end{equation*} where $\operatorname{B}'$ denotes  $\operatorname{B}$ acting on functions with argument $x'$.
\end{lem}
In probability, the mean and covariance functions of a random process under a pushforward are essentially changed according to how the operator $\operatorname{B}$ acts on them. For instance, a GP under a linear operator $\operatorname{B}$ is still a GP with a new mean and covariance function transformed by $\operatorname{B}$. 

\begin{example}
\label{bg}
One common choice for $g \sim GP(0, k)$ is
\begin{align*}
    k(x,x') = \sigma^2 \exp{ \left(-\frac{\|x - x'\|^2}{2l^2}\right), }
\end{align*} where $\sigma^2$ is the variance parameter, and $l$ is the length scale parameter. 
Taking the operator $B = \partial_{x_j}$, the new covariance function of $B_*g$ is calculated as:
\begin{align*}
\frac{\partial^2}{\partial x_j \partial x_j'} k(x, x') 
&= \frac{\sigma^2}{l^2} \exp{\left(-\frac{\|x - x'\|^2}{2l^2} \right)} \nonumber \\
&\quad + \frac{\sigma^2}{l^4} (x_j - x_j')^2 \exp{\left(-\frac{\|x - x'\|^2}{2l^2} \right)}.
\end{align*}
\end{example}
Instead of solving the differential equation, we construct a covariance function which is in the null space of the PDE operator  in both variables. In the case of 2$d$-wave equation, we require $Lk(L')^T=0$ with $L$ as in \eqref{2dwaveeq} (see Lemma~\ref{lemma-1}). The covariance function implicitly encodes the constraints imposed by the differential operator $L$. This ensures that the random functions drawn from the process also satisfy the differential equation.

\begin{figure}[t]
        \centering
        \includegraphics[width=\columnwidth]{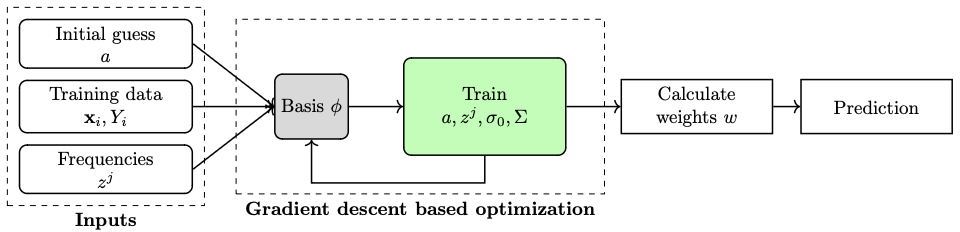}
    
    \caption{Scheme describing our solution to the inverse problem for the wave equation $u_{tt}=a^2(u_{xx}+u_{yy})$ with unknown wave speed $a$.
    We start with an initial guess for $a$, training data $\mathbf x_i{, Y_i}$, and frequencies $  z^j$ which are randomly chosen from a Gaussian distribution, cf. \eqref{epequation}. The basis $\phi$ is constructed based on the characteristic variety \eqref{2dwavecv} of the PDE. For inverse problems, frequencies $z^j$, error term covariance $\sigma_0$ and Gaussian prior covariance $\Sigma$ are trained using gradient based optimization method. For direct problems, we use the same process with one less variable, wave speed $a$. The weights $w$ are calculated using optimal $\hat{z}^j$'s, $\hat{\Sigma}$, and $\hat{\sigma_0}$ by $\hat{w} = \hat{A}^{-1}\hat{\phi}Y$, where $A = \phi\phi^T + \sigma_0^2\Sigma^{-1}$. By \eqref{predf}, our prediction is $\hat{Y} = \phi_*^T\hat{w}$, where $\phi_*$ is the matrix with columns for test data.}
    \label{fig:stacked_process}
\end{figure}

\section{Methods}
Due to their wide range of applications, wave equations will be the main example of this paper. In particular, we will look at  2$d$-wave equations $Lu = 0$ for $L$ in \eqref{2dwaveeq} with constant coefficients. As explained in Section \ref{epsection}, we use the Macaulay2 package \texttt{solvePDE} to see that the characteristic variety \eqref{2dwavecv} is irreducible and Noetherian multipliers are  $D_{i,j}(\mathbf x,z) = 1$.
 From \eqref{epequation},
\begin{equation} \label{ep1}
    f(\mathbf x) = \sum^{m}_{j=1}w^j e^{\langle x,z^j\rangle} \text{ for }\mathbf x=(x,y,t)\in\mathbb R^3,
\end{equation}
where $w^j$'s are the weights and $z^j$'s are the frequencies in the characteristic variety given by \eqref{2dwavecv}.

\cite{harkonen2023gaussian} introduced the (S-)EPGP algorithm, which defines a GP with realizations of the form in \eqref{ep1}, where $z^j$'s are purely imaginary. Equation \eqref{ep1} gives an exact solution for any linear PDE with constant coefficients. In practice, we can approximate the realization $f(\mathbf x)$ by $n$ data points ${(\mathbf x_i,Y_i)}$, $m$ frequency points $z^j=z_{1}^j+\sqrt{-1}z_{2}^j$ and weights $w^j = w_{1}^j + \sqrt{-1}w_{2}^j$ , where $z_{1}^j, z_{2}^j\in\mathbb R^3 \text{ and } w_{1}^j,w_{2}^j \in \mathbb{R}$. The frequencies are initially sampled from a standard Gaussian distribution.
\begin{align} \label{appfx}
    f(\mathbf x) &\approx \sum_{j = i}^m w^je^{\mathbf x\cdot z^j}\\
    &= \sum_{j=1}^m \bigg[e^{z_{1}^j\cdot \mathbf x}(w_{1}^j\cos(z_{2}^j\cdot \mathbf x) - w_{2}^j\sin(z_{2}^j\cdot \mathbf x)) \notag\\
    &\quad + \sqrt{-1}e^{z_{1}^j\cdot \mathbf x}(w_{1}^j\cos(z_{2}^j\cdot \mathbf x) + w_{2}^j\sin(z_{2}^j\cdot \mathbf x))\bigg].\notag
\end{align} 

We are only interested in real-valued solutions of the wave equations. By taking the real part of \eqref{appfx}, the solution is approximated by
\begin{align} \label{predf}
    f(\mathbf x) 
    &\approx \sum_{j=1}^m e^{z_{1}^j\cdot \mathbf x}(w_{1}^j\cos(z_2^j\cdot \mathbf x) - w_{2}^j\sin(z_2^j\cdot \mathbf x)) \notag\\
    &= \phi(\mathbf x)^T w
\end{align}
where $w$ is a matrix with entries of $w_{1}^je^{z_1^j\cdot \mathbf x}$ and $w_{2}^je^{z_1^j\cdot \mathbf x}$, and $\phi(\mathbf x)$ is a matrix with entries of $\cos(z_2^j\cdot \mathbf x)$ and $\sin(z_2^j\cdot \mathbf x)$. In our experiments, we absorbed $e^{z_{1}^j\cdot \mathbf x}$ in the weight $w$.

Instead of using exponential basis $\{e^{\mathbf x_i\cdot z^j}\}$, we use the basis elements $[\cos{(\mathbf x_i\cdot z^j)}, \sin{(\mathbf x_i\cdot z^j)}]$, where $\mathbf x_i$'s are the data points, and $z^j$'s are the real frequencies in the characteristic variety. This reformulation eliminates the need for complex numbers in the algorithm, simplifying the computation process. By operating only with real-valued functions, the method enhances computational efficiency and reduces memory requirements, leading to faster execution.

\begin{table}[tbp]
\centering
\caption{Comparison of RMSE (square root of the average squared differences) and MAE (average of the absolute differences) between direct and inverse problems for the true solution $\cos(x - \sqrt{3}t) + \cos(y - \sqrt{3}t)$ of the wave equation $u_{tt}=a^2(u_{xx}+u_{yy})$. In the direct problem, $a=3$ is known; in the inverse problem, the correct value is learned. \text{n\_pts} is the number of data points; \text{n\_MC} is the number of frequency points. Both predictions are very accurate. Solving the inverse problem requires more resources and is less precise because we need to learn one more parameter; its approximation leads to small errors. 
}
\label{cpdandi}
\begin{tabular}{|c|c|c|c|}
\hline
\multicolumn{4}{|c|}{\textbf{Direct Problem}} \\ \hline
\textbf{n\_pts} & \textbf{n\_MC} & \textbf{RMSE} & \textbf{MAE} \\ \hline
100 & 10 & $7.91\times 10^{-7}$ & $9.20\times 10^{-7}$ \\ \hline
\multicolumn{4}{|c|}{\textbf{Inverse Problem}} \\ \hline
\textbf{n\_pts} & \textbf{n\_MC} & \textbf{RMSE} & \textbf{MAE} \\ \hline
1000 & 100 & $5.63\times 10^{-5}$ & $7.65\times 10^{-5}$ \\ \hline
\end{tabular}
\end{table}

\begin{table*}[]
\caption{Error comparison for different true solutions and corresponding equations. RMSE is particularly sensitive to large errors due to the squaring of residuals. MAE measures the average magnitude of errors between predicted values and actual values, without considering their direction. The results demonstrate that all errors are very small, indicating accurate predictions. Furthermore, the parameter $a$, which we optimize for, is closely approximating the true values $a^2=3,\,1.5,\,3$ respectively in each case, confirming the reliability of the proposed method.}
\label{tab:error_comparison}
\begin{tabular}{l|l|r|r|r|r|r}
\toprule
\multicolumn{1}{c}{\multirow{2}{*}{True Solution $u(x,y,t)$}} & \multicolumn{1}{c}{\multirow{2}{*}{Corresponding Equation}} & \multicolumn{2}{c}{Direct}                                  & \multicolumn{2}{c}{Inverse}                                 \\ \cline{3-4} \cline{5-7}
\multicolumn{1}{c}{}                                            & \multicolumn{1}{c}{}                                        & \multicolumn{1}{c}{RMSE} & \multicolumn{1}{c}{MAE} & \multicolumn{1}{c}{RMES} & \multicolumn{1}{c}{MAE}& \multicolumn{1}{c}{$\bm{a^2}$} \\
\midrule
$\cos(x - \sqrt{3}t) + \cos(y - \sqrt{3}t)$                     & $u_{tt} = 3(u_{xx} + u_{yy})$                         & $3.067\times 10^{-8}$                            & $1.065\times 10^{-8}$                            & $5.632\times 10^{-5}$                           & $7.647\times10^{-5}$                         &\textbf{3.0002} \\ 
$(x+y - \sqrt{3}t)^2$                                           & $u_{tt} = 1.5(u_{xx} + u_{yy})$                       & $3.459\times 10^{-4}$                            & $9.410\times 10^{-5}$                            & $3.006\times10^{-4}$                            & $1\times10^{-4}$                           &\textbf{1.5018} \\
$\cos(3(x - \sqrt{3}t)) + \cos(6(y - \sqrt{3}t))$               & $u_{tt} = 3(u_{xx} + u_{yy})$                         & $2.483\times10^{-7}$                            & $2.099\times10^{-7}$                              & $3.508\times10^{-5}$                             &$1.744\times10^{-5}$   & \textbf{2.9999}   \\                      
\bottomrule
\end{tabular}
\end{table*}

\begin{figure*}[h!] 
    \centering
    \begin{subfigure}[b]{0.24\textwidth} 
        \includegraphics[width=\textwidth]{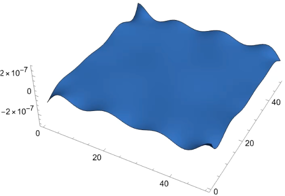}
        \caption{{Direct for \eqref{truea3}}}
        \label{fig:image1}
    \end{subfigure}
    \hfill
    \begin{subfigure}[b]{0.24\textwidth}
        \includegraphics[width=\textwidth]{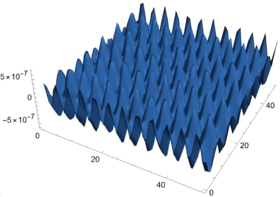}
        \caption{{Direct for \eqref{truey36}}}
        \label{fig:image2}
    \end{subfigure}
    \hfill
    \begin{subfigure}[b]{0.24\textwidth}
        \includegraphics[width=\textwidth]{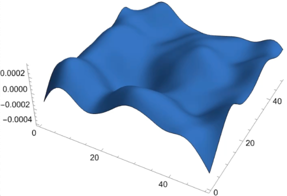}
        \caption{{Inverse for \eqref{truea3}}}
        \label{fig:image3}
    \end{subfigure}
    \hfill
    \begin{subfigure}[b]{0.24\textwidth}
        \includegraphics[width=\textwidth]{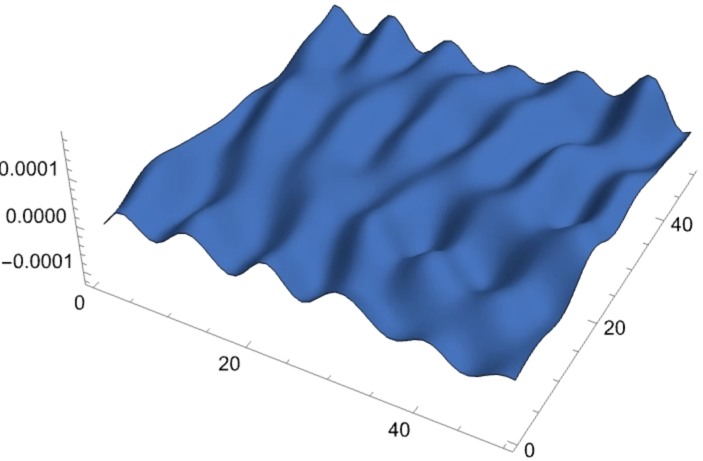}
        \caption{{Inverse for \eqref{truey36}}}
        \label{fig:image4}
    \end{subfigure}
    \caption{These four images show the difference between the prediction of our solution to the wave equation $u_{tt}=a^2(u_{xx}+u_{yy})$ and the true solution with unknown true wave speed $a = \sqrt{3}$. We  consider different true solutions $u(x,y,t)$. Fig. (\ref{fig:image1}) and Fig. (\ref{fig:image3}) concern the direct and inverse problem respectively with true solution $u(x,y,t) = \cos(x - \sqrt{3}t)+\cos(y - \sqrt{3}t)$. Fig.  (\ref{fig:image2}) and Fig. (\ref{fig:image4}) concern the  direct and inverse problem respectively with true solution $u(x,y,t) = \cos(3(x - \sqrt{3}t))+\cos(6(y - \sqrt{3}t))$. The errors in the all these cases are very small -- at most of the order of $10^{-4}$, indicating highly accurate predictions. 
    }
    \label{fig:four_images} 
\end{figure*}

\subsection{Loss function}
To turn the approximated solution $\{f(\mathbf x)\}$ into a GP, let $w \sim N(0, \Sigma)$, where $\Sigma$ is a diagonal matrix with positive entries $\sigma_j^2$, $j = 1,\ldots,m$. Then by Lemma~\ref{lemma-1}, the covariance function is of the form 
\begin{align*}
    k_{\text{EPGP}}(\mathbf x,\mathbf x') = \phi(\mathbf x)^T\Sigma \phi(\mathbf x')
\end{align*}
where $\phi$ is the basis matrix evaluated at different data points $\mathbf x$ and $\mathbf x'$.
Suppose that $Y = f(\mathbf{X}) + \epsilon$, where $\epsilon \sim N(0, \sigma_0^2\mathbf{I}_n)$, so the marginal likelihood $P(Y|\mathbf{X})$ is obtained by integrating out $w$ from the joint distribution $P(Y|\mathbf{X},w)P(w)$. The negative log-marginal likelihood is
\begin{align} \label{nlml}
    \text{NLML} &= - \frac{1}{2\sigma_0^2}\left(Y^T Y - Y^T \phi^T A^{-1} \phi Y\right) \notag\\
    &\quad - \frac{n - m}{2} \log{\sigma_0^2} - \frac{1}{2} \log{|\Sigma|} - \frac{1}{2} \log{|A|},
\end{align} 
see \citep{harkonen2023gaussian}. NLML will serve as the loss function for the optimization process in this paper. Here, $Y_i$ is the observed value corresponding to our prediction  $\phi(\mathbf x_i)^Tw$,
$\mathbf{X}$ is the matrix of all $n$ data points $\mathbf x_i$, and 
\begin{align} \label{covmatrixA}
    A = \phi\phi^T + \sigma_0^2\Sigma^{-1}.
\end{align}
The term $\sigma_0^2\Sigma^{-1}$ captures the information from the prior distribution and also ensures that $A$ is  positive definite. 

During the training process, $a$, $\sigma_0^2$, $z^j$'s and $\Sigma$ are learned using a gradient descent-based optimization algorithm with loss function \eqref{nlml}. After the optimal $\hat a$, $\hat{\sigma}_0^2$, $\hat{z}^j$'s and $\hat{\Sigma}$ are obtained, the weight $\hat{w}$ is calculated as 
\begin{align*}
    \hat{w} = \hat{A}^{-1}\hat{\phi}Y.
\end{align*}

An issue here is to efficiently calculate the inverse of the covariance matrix $A$. We use Cholesky decomposition $\hat A = LL^T$ to find $\hat A^{-1}$ and the prediction is given by
\begin{align*}
\phi_*^T\hat{w} = \phi_*^T \hat A^{-1} \hat \phi Y= \phi_*^T (L)^{-1} (L^T)^{-1} \hat \phi Y, 
\end{align*}
where $\phi_*$ is the matrix with columns for test data, and $L$ is the lower triangular matrix such that $\hat A^{-1} = (L)^{-1} (L^T)^{-1}$.

\begin{figure}[b]
\begin{center}
\includegraphics[width=\columnwidth]{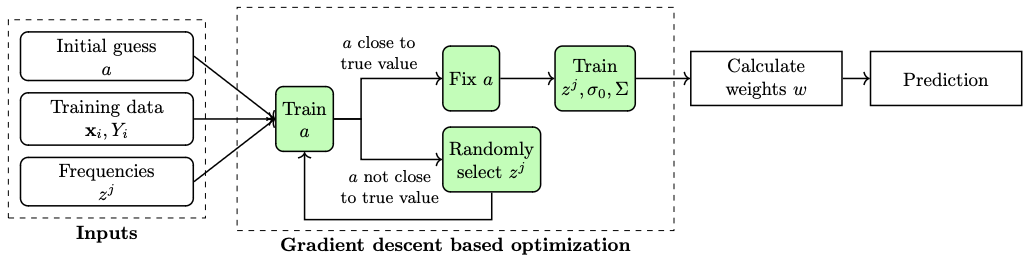}
\caption{
Scheme describing our solution to the inverse problem for the wave equation when the estimation of the wave speed $a$ gets stuck in a local minimum. This issue arose when considering the solution $ u(x,y,t) = \cos(3(x - \sqrt{3}t))+\cos(6(y - \sqrt{(3}t))$, which requires fitting higher frequencies.
To address this issue, instead of training $a$, $z^j$, $\sigma_0^2$ and $\Sigma$ simultaneously, we train the wave speed $a$ first. If the difference between squared $a$ and true wave speed is less than $10^{-6}$, then we proceed to optimize the remaining parameters $z^j$, $\sigma_0^2$, $\Sigma$. Otherwise, the process is restarted by randomly selecting new frequencies $z^j$'s to improve initialization.
In other examples, the  scheme in Fig.~\ref{fig:stacked_process} sufficed.
}
\label{fig:invhighfreq}
\end{center}
\end{figure}

\begin{figure*}[h] 
    \centering
    \begin{subfigure}[b]{0.23\textwidth} 
        \includegraphics[width=\textwidth]{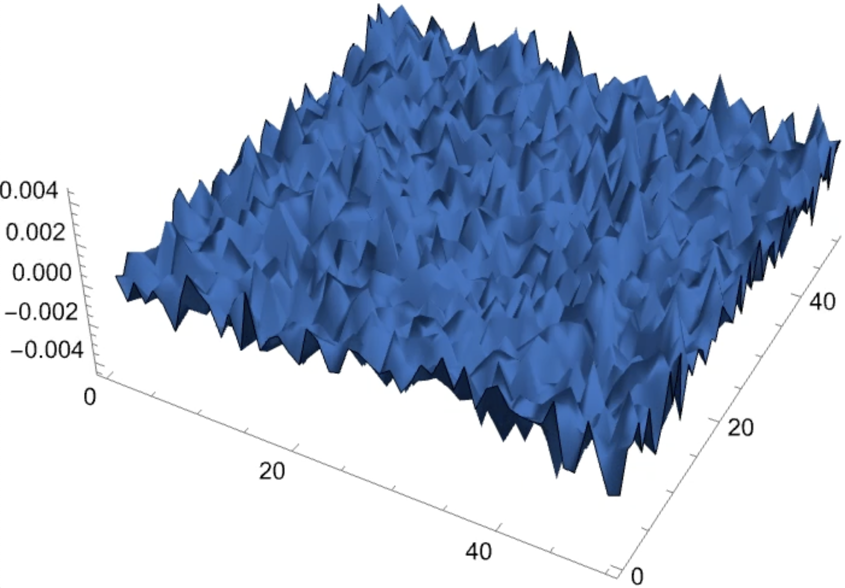}
        \caption{{Direct for \eqref{truea3} with Noise}}
        \label{fig:imagen1}
    \end{subfigure}
    \hfill
    \begin{subfigure}[b]{0.23\textwidth}
        \includegraphics[width=\textwidth]{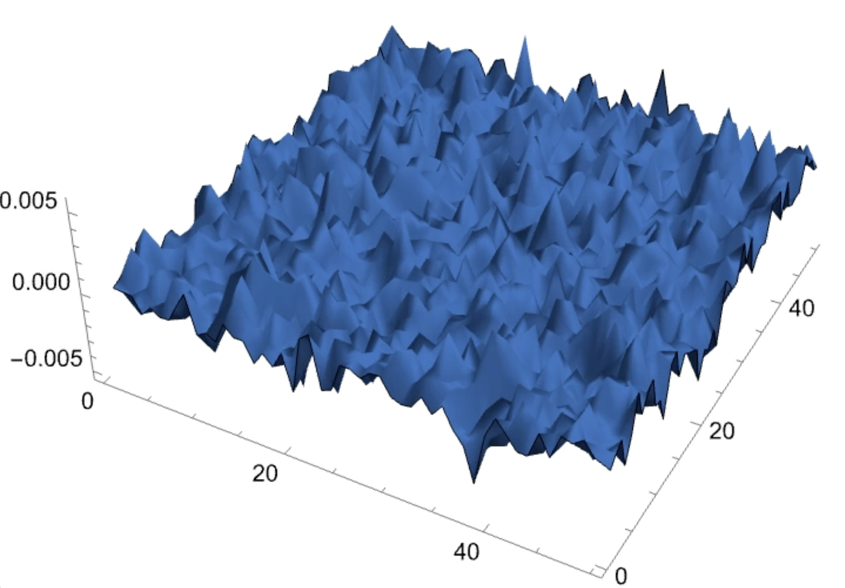}
        \caption{{Direct for \eqref{truey36} with Noise}}
        \label{fig:imagen2}
    \end{subfigure}
    \hfill
    \begin{subfigure}[b]{0.23\textwidth}
        \includegraphics[width=\textwidth]{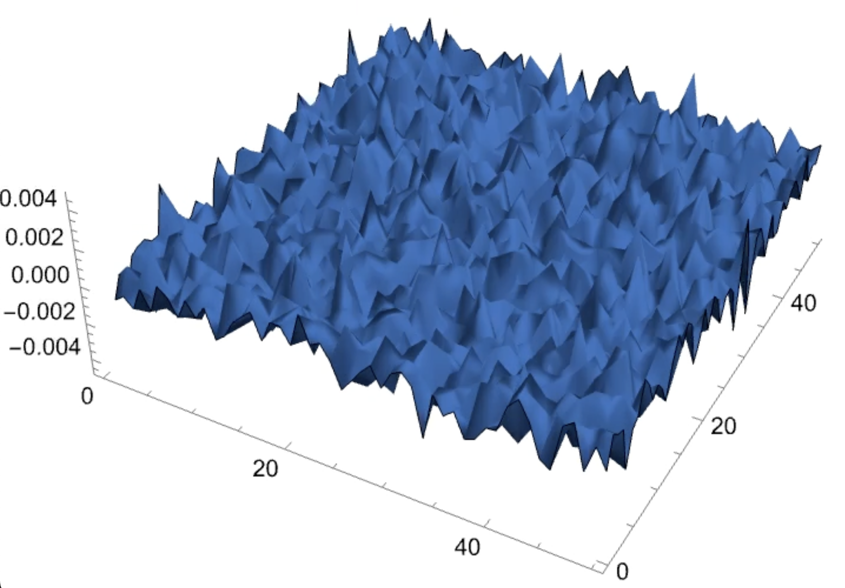}
        \caption{{Inverse for \eqref{truea3} with Noise}}
        \label{fig:imagen3}
    \end{subfigure}
    \hfill
    \begin{subfigure}[b]{0.23\textwidth}
        \includegraphics[width=\textwidth]{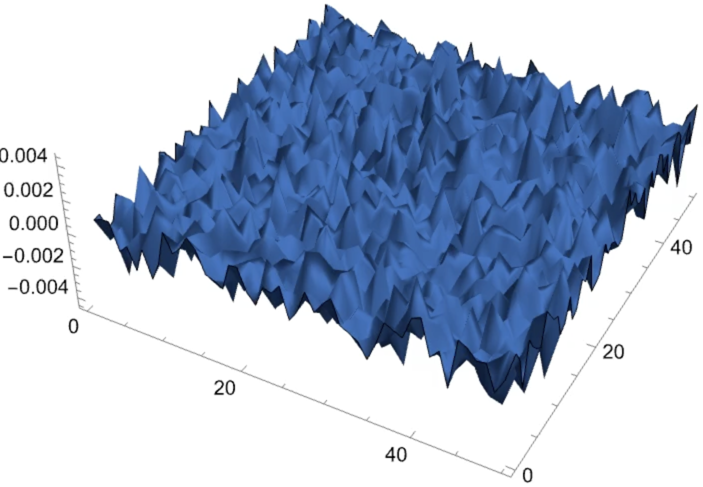}
        \caption{{Inverse for \eqref{truey36} with Noise}}
        \label{fig:imagen4}
    \end{subfigure}
    \caption{These four images show the difference between the prediction and the true solution to the wave equation $u_{tt}=a^2(u_{xx}+u_{yy})$ for unknown wave speed $a = \sqrt{3}$. As data, we considered true solution $u(x,y,t)$ to which we added Gaussian noise. Figure (\ref{fig:image1}) and (\ref{fig:image3}) concern the direct and inverse problem respectively with true solution $u(x,y,t) = \cos(x - \sqrt{3}t)+\cos(y - \sqrt{3}t)$. Figure (\ref{fig:image2}) and (\ref{fig:image4}) concern the direct and inverse problem respectively with true solution $u(x,y,t) = \cos(3(x - \sqrt{3}t))+\cos(6(y - \sqrt{3}t))$. Compared to the noise-free experiments, the errors here are relatively larger, of the order of $10^{-3}$ compared with $10^{-4}$. This was to be expected due to the presence of noise. The errors are sufficiently small to ensure accurate predictions and primarily reflect the added noise rather than inaccuracies in the prediction method.}
    \label{fig:four_images_noise}
\end{figure*}

\subsection{Direct problem vs. inverse problem}
For the direct problem, i.e., given wave speed $a$, we are going to learn the parameters $\hat{\theta} = (\hat{\sigma}_0^2, \hat{z}^j, \hat{\Sigma})$ so that we can reconstruct the wave from the learned frequency. For the inverse problem, in addition to learning these parameters, we learn the wave speed $a$ from the sample points randomly selected from a true solution. The  covariance matrix $A$ calcuated in \eqref{covmatrixA} acts as a prior that constrains the solution to satisfy the PDEs. 

The results presented in Table~\ref{cpdandi} demonstrate that even if the number of data points and frequencies is both increased by 10 times, the performance of the inverse problem cannot be as good as that of the direct problem. Nevertheless, for the inverse problem in Table~\ref{cpdandi}, the squared wave speed $a^2$ is correctly learned as $3.0002$, closely approximating the true value $a^2 = 3$. The direct problem is solved with remarkable speed, achieving an accuracy of $10^{-8}$ in just 30 seconds. In contrast, the inverse problem requires approximately 15 minutes to reach a similar level of accuracy. The increase in error is due to the inexact value of $a$ during the iteration.

Although the processes for solving the direct and inverse problems differ only by  one  parameter $a$, the results in Table~\ref{cpdandi} indicate challenges in achieving comparable levels of accuracy and computational efficiency for the inverse problem. Further details and specialized techniques for obtaining better results of the inverse problems are presented in the following section.

\begin{table*}[t]
\caption{Error comparison for different true solutions and corresponding equations with Gaussian noise. Compared to results of noise-free experiments, the errors here are relatively larger due to the presence of noise. The parameter $a$ is estimated correctly through optimization. These results indicate that the proposed approach exhibits noise tolerance and maintains high accuracy.}
\label{tab:error_comparison_noise}
\begin{tabular}{l|l|r|r|r|r|r}
\toprule
\multicolumn{1}{c}{\multirow{2}{*}{True Solution $u(x,y,t)$}} & \multicolumn{1}{c}{\multirow{2}{*}{Corresponding Equation}} & \multicolumn{2}{c}{Direct}                                  & \multicolumn{2}{c}{Inverse}                                 \\ \cline{3-4} \cline{5-7}
\multicolumn{1}{c}{}                                            & \multicolumn{1}{c}{}                                        & \multicolumn{1}{c}{RMSE} & \multicolumn{1}{c}{MAE} & \multicolumn{1}{c}{RMES} & \multicolumn{1}{c}{MAE}& \multicolumn{1}{c}{$\bm{a^2}$} \\
\midrule
$\cos(x - \sqrt{3}t) + \cos(y - \sqrt{3}t)$                     & $u_{tt} = 3(u_{xx} + u_{yy})$                         & $9.868\times 10^{-4}$                            & $8\times 10^{-4}$                            & $9.812\times 10^{-4}$                           & $8\times10^{-4}$                         &\textbf{2.99989} \\ 
$\cos(3(x - \sqrt{3}t)) + \cos(6(y - \sqrt{3}t))$               & $u_{tt} = 3(u_{xx} + u_{yy})$                         & $8.254\times10^{-4}$                            & $9\times10^{-4}$                              & $1.095\times10^{-3}$                             &$8\times10^{-4}$   & \textbf{2.99999}   \\                      
\bottomrule
\end{tabular}
\end{table*}

\section{Example and Experiments}
In this section, we illustrate the algorithm by testing both direct and inverse problems for the 2$d$-wave equation $u_{tt} = a^2(u_{xx} + u_{yy})$ with different types of true solutions. The Noetherian multiplier of 2$d$-wave equation is 1. The training grid is set as $[-6,6]\times[-6,6]\times[0,12]$ for spatial and time variables. For the training process, we start by randomly choosing $n = 10,000$ data points $\mathbf x_i=(x_i, y_i, t_i)$ from the grid and $m = 1000$ frequency points $z^j=(z^j_1,z^j_2)$ from a standard normal distribution. For each experiment, we use a true solution $u$ and sample values $Y_i=u(\mathbf x_i)$ or $Y_i= u(\mathbf x_i)+\varepsilon_i$, where $\varepsilon_i$ is Gaussian noise.

\subsection{2$d$-wave equations with high frequency solutions}
With $\mathbf{x}_i = (x_i,y_i,t_i)$, $i = 1,2,\ldots,n$, as the data points and  ${z}^j = (z_1^j,z_2^j,z_3^j)$, $j=1,\ldots m$, as the frequencies, the solutions to the 2$d$-wave equation of the form $e^{x_iz_1^j + y_iz_2^j+t_iz_3^j}$ are subject to the characteristic equation
\begin{align*} 
    ({z_3^j})^2 = a^2\left(({z_1^j})^2 + ({z_2^j})^2\right),\quad j = 1,\ldots,m
\end{align*}
and have a $4m \times n$ basis matrix $\phi$ with entries 
\begin{align*}
    \cos{\left(x_iz_1^j + y_iz_2^j \pm t_i\sqrt{a^2(({z_1^j})^2 + ({z_2^j})^2)}\right)} \notag\\
    \sin{\left(x_iz_1^j + y_iz_2^j \pm t_i\sqrt{a^2(({z_1^j})^2 + ({z_2^j})^2)}\right)}.
\end{align*}


For low-frequency solutions, characterized by fewer oscillations, the optimization process adheres precisely to the steps outlined in Figure~\ref{fig:stacked_process}. In case of high-frequency solutions, the procedure for solving direct problems remains consistent with Figure~\ref{fig:stacked_process}. However, it is not sufficient for the inverse problem as the wave speed $a$ was trapped in an incorrect value during the optimization process. This situation indicates the possibility of inverse problems converging to local minima, thereby affecting the accuracy.

To address this, we adjusted our training approach. Instead of simultaneously optimizing $a$, $z^j$' s, $\Sigma$ and $\sigma_0^2$, we adopted a sequential strategy. Initially, we randomly selected $z^j$' s, then only trained $a$ until it closely approximated the true value. Once $a$ is close  to the true value, we proceeded to train $z^j$'s, $\Sigma$ and $\sigma_0^2$. The detailed process is shown in Figure~\ref{fig:invhighfreq}. 

\subsection{Experiments with different true solutions}
For one of the true low-frequency solutions 
\begin{align}\label{truea3}
    u(x,y,t) = \cos(x - \sqrt{3}t)+\cos(y - \sqrt{3}t),
\end{align}
which corresponds to a wave speed $a = \sqrt{3}$, highly accurate results are obtained, as presented in Table~\ref{tab:error_comparison}, for both direct and inverse problems, starting from $a = 1$, while experiments with other initial guesses worked equally well, e.g. $a=10$. 
The root mean square error (RMSE) and mean absolute error (MAE) are on the order $10^{-8}$ for direct problems and of the order $10^{-5}$ for inverse problems. These low errors demonstrate that the predicted solution closely approximates the true solution \eqref{truea3}, highlighting the effectiveness of the approach in capturing wave propagation.

We also explored  a high-frequency solution 
\begin{align}\label{truey36}
    u(x,y,t) = \cos(3(x - \sqrt{3}t))+\cos(6(y - \sqrt{(3}t)).
\end{align}
The direct problem is effectively solved, achieving both RMSE and MAE on the order of $10^{-7}$, as shown in Table~\ref{tab:error_comparison}. A satisfactory result for the inverse problem is obtained when the squared wave speed is initialized at $a^2 = 2$. The results presented in Table~\ref{tab:error_comparison} are derived based on the procedure shown in Figure~\ref{fig:invhighfreq}. As observed in Table~\ref{tab:error_comparison}, the errors for direct problems are approximately on the order $10^{-7}$, whereas for inverse problems, the errors are around $10^{-4}$. It is expected that the inverse problem exhibits larger errors due to one more unknown parameter and sensitivity to initial guess. Despite this, the method demonstrates robust performance in reconstructing the wave propagation and learning the correct wave speed. 

Even after adding Gaussian noise to the true solutions \eqref{truea3} and \eqref{truey36},  the results remain accurate, as shown in Table~\ref{tab:error_comparison_noise} and Figure \ref{fig:four_images_noise}. Compared to experiments without noise, the difference between the prediction and true solutions increases significantly, approximately 10,000 times for direct problems and 10 times for inverse problems. However, these discrepancies are still small relative to the high accuracy of the method, since the performance in the noiseless experiments was exceptionally precise, underscoring the robustness of the approach even in the presence of noise.

We also tried
$
    u(x,y,t) = (x + y - \sqrt{3}t)^2
$
which is the solution for the wave speed $a = \sqrt{1.5}$. Results are  shown in Table~\ref{tab:error_comparison} with 1000 data points and 100 frequency points. 

\section{Conclusion}
This paper presents a computationally efficient approach for solving inverse problems governed by linear partial differential equations (PDEs) using Gaussian Process (GP) regression. By using algebraically informed priors derived from commutative algebra and algebraic analysis, the proposed method ensures both accuracy and interpretability while maintaining computational efficiency. The implementation of priors using the computer algebra software Macaulay2 enables effective handling of complex relationships in PDE solutions.

Through extensive experiments on 2$d$ wave equations, the method demonstrates robust performance in learning wave speed parameters and accurately reconstructing wave propagation even in the presence of  noise. The results show that while the direct problem achieves high precision with minimal computational cost, the proposed approach effectively mitigates the challenges for the inverse problems, i.e., sensitivity to initial conditions and presence of local minima, achieving satisfactory accuracy. 

The strength of our method is that it is implementable for any system of linear PDE with constant coefficients. In experiments that will be published in forthcoming work, we observed that the method works equally to identify parameters in elliptic and parabolic equations, e.g. the thermal diffusivity in heat equation. We aim to extend our method for problems with variable coefficients.

Other strengths of our method is that it is a meshless method of unsupervised learning which can be adapted to fit both boundary and initial conditions.

One significant direction that is well in reach is the application of our method to solve vectorial wave equations.  Practical examples  include applications to Maxwell's equations in electromagnetism and the seismic inversion problem; see \cite{symes2009seismic}. Here one would identify the stress tensor in the elasticity equations. Future studies could focus on adapting the GP framework to handle large-scale seismic datasets and improve subsurface structure estimation. 
\bibliography{ref}             
\end{document}